\def\paperID{0000}
\def\confName{CVPR}
\def\confYear{2024}

\def\paperTitle{LGmap: Local-to-Global Mapping Network for Online Long-Range Vectorized HD Map Construction
}

\def\authorBlock{
    Kuang Wu    \thanks{Equal Contribution} \qquad
    Sulei Nian  \footnotemark[1] \qquad
    Can Shen    \qquad
    Chuan Yang  \thanks{Tech Lead} \qquad
    Zhanbin Li  \thanks{Corresponding Authors} \qquad   \\
    Langge Technology \\
    {\tt\small \{Kuang.Wu, Sulei.Nian, Can.Shen1, Chuan.Yang3, Zhanbin.Li\}@geely.com}
}

\newif\ifreview 
\newif\ifarxiv 
\newif\ifcamera \newcommand{\cameraready}{\cameratrue}
\newif\ifrebuttal 

\cameraready 

\pdfoutput=1
\documentclass[10pt,twocolumn,letterpaper]{article}
\ifreview \usepackage[review]{cvpr} \fi
\ifarxiv \usepackage[pagenumbers]{cvpr} \fi
\ifrebuttal \usepackage[rebuttal]{cvpr} \fi
\ifcamera \usepackage{cvpr} \fi

\usepackage{graphicx}	
\usepackage{amsmath}	
\usepackage{amssymb}	
\usepackage{booktabs}
\usepackage{times}
\usepackage{microtype}
\usepackage{epsfig}
\usepackage[table,xcdraw,dvipsnames]{xcolor}
\usepackage{caption}
\usepackage{float}
\usepackage{placeins}
\usepackage{color, colortbl}
\usepackage{stfloats}
\usepackage{enumitem}
\usepackage{tabularx}
\usepackage{xstring}
\usepackage{multirow}
\usepackage{xspace}
\usepackage{url}
\usepackage{subcaption}
\usepackage{xcolor}
\usepackage[hang,flushmargin,para]{footmisc}

\ifcamera \usepackage[accsupp]{axessibility} \fi

% ...

\ifarxiv  \fi

\newcommand{\R}[1]{{%
    \textbf{%
        \ifstrequal{#1}{1}{\textcolor{red}{R#1}}{%
        \ifstrequal{#1}{2}{\textcolor{blue}{R#1}}{%
        \ifstrequal{#1}{3}{\textcolor{magenta}{R#1}}{%
        \ifstrequal{#1}{4}{\textcolor{teal}{R#1}}{%
                           \textcolor{cyan}{R#1}%
        }}}}%
    }%
}}

\usepackage{xr-hyper}

\makeatletter
\newcommand*{\addFileDependency}[1]{
  \typeout{(#1)}
  \@addtofilelist{#1}
  \IfFileExists{#1}{}{\typeout{No file #1.}}
}

\makeatother
\newcommand*{\myexternaldocument}[1]{
    \externaldocument{#1}
    \addFileDependency{#1.tex}
    \addFileDependency{#1.aux}
}

\definecolor{cvprblue}{rgb}{0.21,0.49,0.74}
\usepackage[pagebackref,breaklinks,colorlinks,citecolor=cvprblue]{hyperref}
\usepackage[capitalize]{cleveref}
\crefname{section}{Sec.}{Secs.}
\crefname{table}{Table}{Tables}
\crefname{figure}{Fig.}{Figs.}

\ifarxiv \crefname{appendix}{App.}{Apps.}
\else \crefname{appendix}{Suppl.}{Suppls.} \fi

\unless\ifarxiv \myexternaldocument{_supplementary} \fi

\begin{document}
\title{\paperTitle}
\author{\authorBlock}
\maketitle

\begin{abstract}
    This report introduces the first-place winning solution for the Autonomous Grand Challenge 2024 - Mapless Driving \cite{wang2024openlane}. In this report, we introduce a novel online mapping pipeline LGmap, which adept at long-range temporal model. Firstly, we propose symmetric view transformation(SVT), a hybrid view transformation module. Our approach overcomes the limitations of forward sparse feature representation and utilizing depth perception and SD prior information. Secondly, we propose hierarchical temporal fusion(HTF) module. It employs temporal information from local to global, which empowers the construction of long-range HD map with high stability. Lastly, we propose a novel ped-crossing resampling. The simplified ped crossing representation accelerates the instance attention based decoder convergence performance. Our method achieves 0.66 UniScore in the Mapless Driving OpenLaneV2 test set.
\end{abstract}
\section{Introduction}
\label{sec:intro}


The High-Definition (HD) map is designed for high-precision autonomous driving. It contains instance-level vectorized representation such as pedestrian crossing, lane divider, road boundaries, etc. The rich semantic information of road topology and traffic rules is important for the navigation of autonomous driving. The Mapless Driving Track \cite{xuyang2022lanesegnet} aims to dynamically construct a local HD map from the images of the surrounding camera on board and the SD map. In this work, we present a multi-stage framework, which decouples the 2D / 3D elements detection and topology prediction tasks. \\
    Our method focuses mainly on three aspects to handle the competition.
\begin{enumerate}
\item Fusion from close to distant. We propose an innovative approach that incorporate both forward projection and backward projection strategies together with SD-map fusion and depth supervision.
\item Fusion from local to global. We present an novel online mapping pipeline adept at both short-range and long-range, which integrates both streaming strategy and stacking strategy.
\item Ped crossing resampling. We simplify the ped crossing to 4 corners, and then uniformly sample 6 points on each edge.
\end{enumerate}

\section{Method}
\label{sec:method}

\begin{figure*}[t]
    \begin{center}
    \includegraphics[width=1.7 \columnwidth]{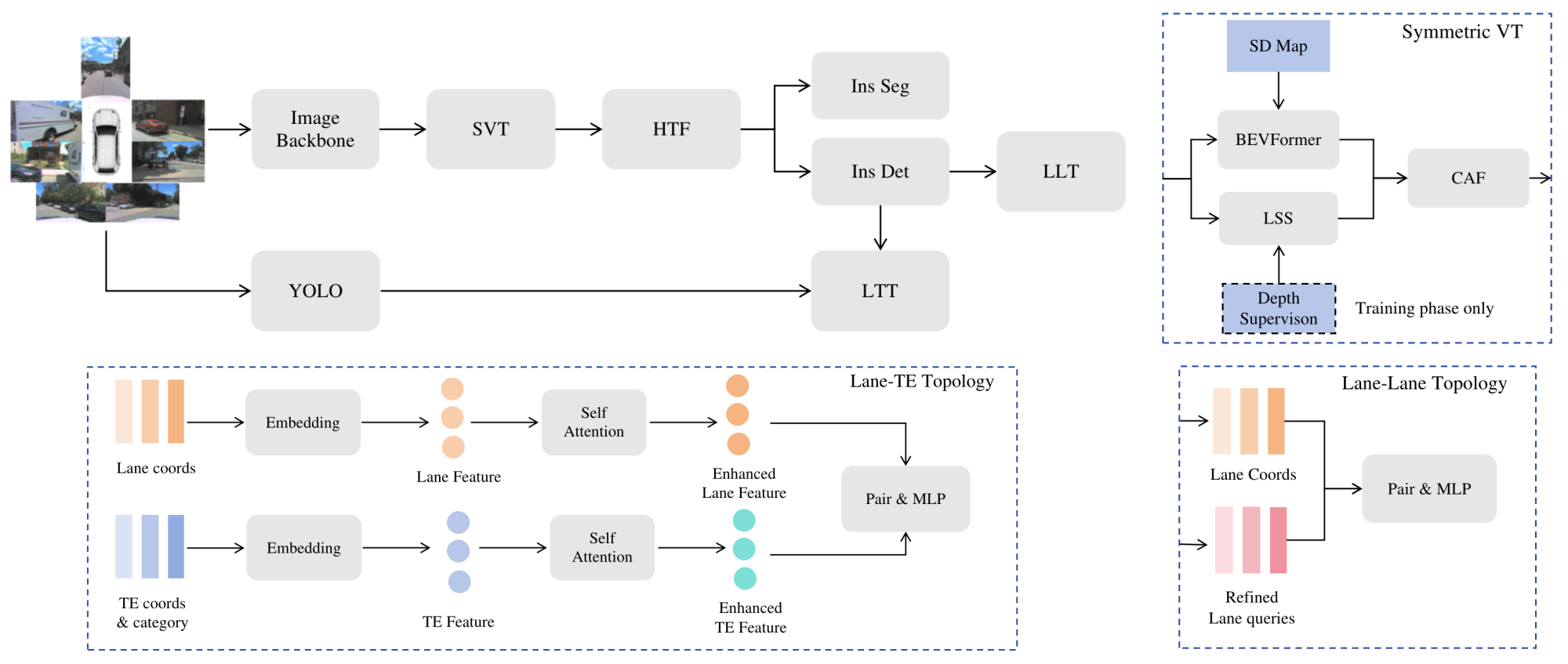}
    \caption{The overall model architecture of LGmap. The entire model is consists of mainly six components: a image backbone equipped with SVT(Symmetric View Transformation), a hierarchical temporal fusion(HTF) module, a unified instance detection and segmentation predictor, a traffic elements detector(YOLO \cite{wang2024yolov9}), a Lane-Lane Topology(LLT) and a Lane-TE Topology(LTT).}
    \label{fig:pipeline}
    \end{center}
    \end{figure*}

This section introduces the details of our method. 
We first introduce the main pipeline of the LGmap architecture, as shown in Fig. \ref{fig:pipeline}. Then the area components and lane segment components are presented. Furthermore, we introduce the traffic elements. Finally, we describe the attention-based heads for topology reasoning.

\subsection{Pipeline}

    \subsubsection{Encoder}
    There are mainly two types of view transformation, forward projection and backward projection. Lift-Splat-Shoot (LSS)\cite{philion2020lift} takes advantage of the depth distribution to model the uncertainty of each pixel’s depth. But the drawbacks of forward projection is discrete and sparse BEV representation. BEVFormer \cite{li2022bevformer} projects 3D points back onto 2D images. As a backward projection, one limitation of BEVFormer is false correlation between 3D and 2D space due to occlusion.
To address these issues, we introduce a symmetric view transformation. The depth-map of each camera is generated from synchronized lidar point cloud. The LSS utilize depth supervision only at the training phase. Given the SD map of the scene, we evenly sample along each of the polylines for a fixed number of points. With sinusoidal embedding, BEVFormer apply cross-attention between the SD map feature representation with features from vision inputs on each encoder layer. In order to fuse BEV representations, we use the channel-attention-based fusion module.

    \subsubsection{Decoder}
    In order to handle different map elements with distinct shape priors, we extend the instance-wise detection decoder with additional segmentation tasks. The unified transformer-based decoder for instance detection and segmentation benefits from both pixel-level classification task and region-level regression task. Additional segmentation branches accelerate the convergence performance of the instance-wise feature embedding.

    \subsubsection{Temporal fusion}
\begin{figure}[t]
    \begin{center}
    \includegraphics[width=0.8 \columnwidth]{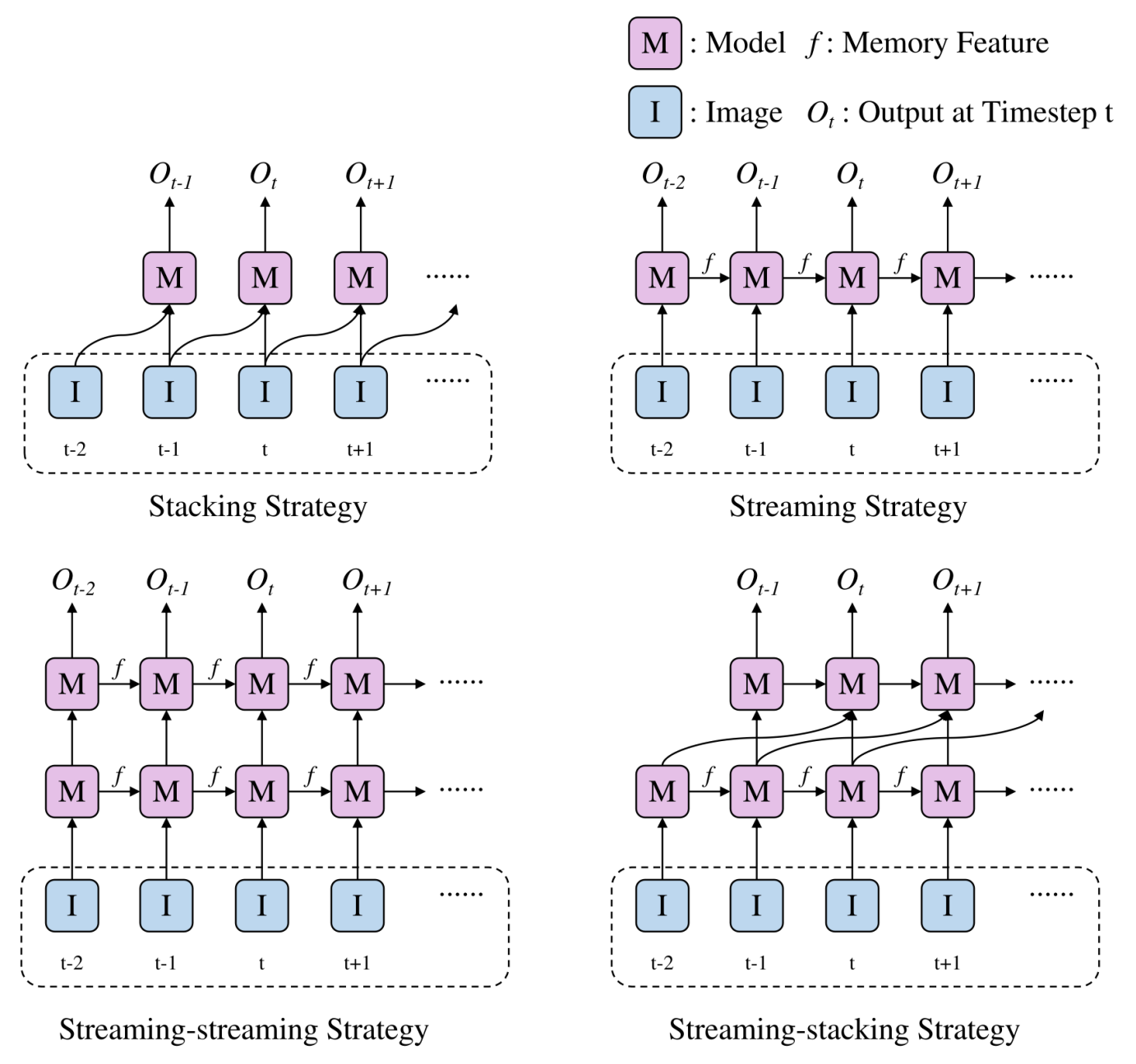}
    \caption{Stacking strategy and streaming strategy are same as StreamMapNet's \cite{yuan2024streammapnet} summary. In order to demonstrate the effectiveness of long-range stacking for streaming-stacking strategy in figure, the stacking previous frame interval parameter is set to 2. Stacking strategy only fuses one previous frame in this figure, and actually it may fuse more than one frame.}
    \label{fig:steamstack}
    \end{center}
    \end{figure}

    The streaming strategy facilitates longer temporal association as the propagated hidden states encode all historical information. But a temporal fuser such as convGRU \cite{chung2014empirical} may still face the problem of forgetting.
The stacking strategy may integrate features from specific previous frames, offering flexibility in fusion of long-range information. The computational cost is linearly related to the number of fused frames.
We propose a novel hierarchical temporal fusion (HTF). The hierarchical temporal fusion fully leverage local fusion capability of streaming strategy and long-range fusion capability of stacking strategy. And it minimizes memory and latency costs compared to the stacking strategy. Here we present two variants of HTF, streaming-streaming strategy and streaming-stacking strategy, as shown in Fig. \ref{fig:steamstack}. For streaming-stacking strategy, we random select N frames from the latest M previous frames for the stacking mode layer during the training phase. And select N frames by a certain distance strides during testing phase.

    \subsubsection{Loss functions}
    Firstly, we adopt classification loss, point2point loss and edge direction loss same as MapTR \cite{liao2022maptr}. 
Secondly, we adopt image segmentation auxiliary dense prediction loss and depth prediction loss same as MapTRv2 \cite{liao2023maptrv2}. 
Thirdly, we adopt BEV instance segmentation loss.
Lastly, we adopt geometric 3D loss. Unlike the geometric loss of GeMap \cite{zhang2023online}, which ignores the Z-axis, we extend the euclidean loss dimension from 2d to 3d.

    \subsection{Area}
\begin{figure}[t]
    \begin{center}
    \includegraphics[width=1.0 \columnwidth]{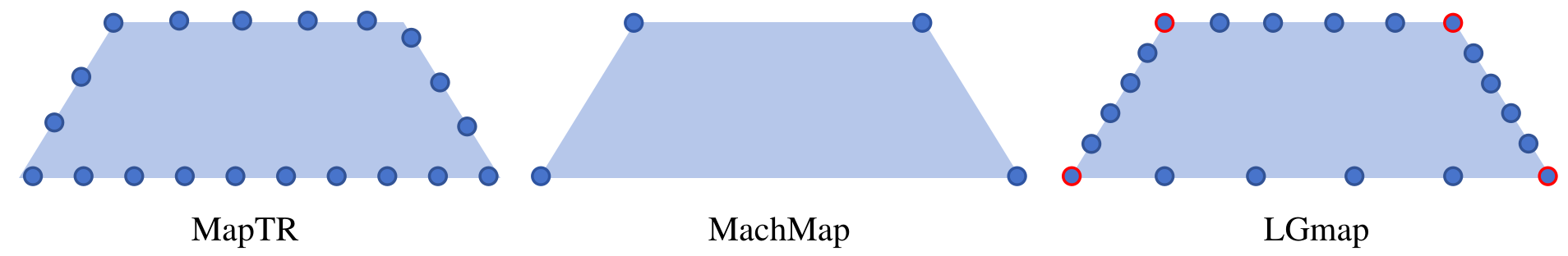}
    \caption{The ped crossing form of MapTR, MachMap and LGmap.
}
    \label{fig:ped}
    \end{center}
    \end{figure}

    Inspired by Machmap \cite{qiao2023machmap}, we simplified the ped crossing by four corners. Then we unified the four corners into the MapTR form of N points. The main difference is that MapTR uses 20 evenly sampled points, MachMap uses 4 points, and we use 6 points evenly sampled on each edge, as shown in Fig. \ref{fig:ped}.

    Our ped-crossing representation keeps four corners as key points, which are essential shape priors. What's more, the permutations of ped crossing are simpler than MapTR. Compared to MapTR's 40 equivalent permutations of one 20 points polygon, LGmap only requires 8. Instead of point-wise permutations, we only use corner-wise permutations. Lastly, preserving corners is beneficial for instance query embedding.

    \subsection{Lane segments}
    Based on the centerline output of the regression branch, an offset branch is introduced to predict the offset to left and right lane boundaries, and two classification branches are introduced to predict the attribute of lane boundary, with reference to LaneSegNet \cite{xuyang2022lanesegnet}.

    \subsection{Traffic elements}
    We utilize YOLOv8 as a base 2D detector, and we utilize YOLOv9 \cite{wang2024yolov9} additionally for model ensemble. Based on the OpenlaneV2 dataset, we propose a series of data augmentation excluding HSV and horizontal flipping, since these tricks may lead to confusion of traffic lights and the direction of traffic signs. The distribution of categories in dataset is highly imbalanced, some categories differing by an order of magnitude. Moreover, pseudo-labels, which are generated on the test set, improve the results. We adopt test-time augmentation(TTA) with the scale range between 0.7-1.4 to improve both the recall of small-objects and large-objects.

    \subsection{Lane-Lane topology}
    We use the TopoMLP method \cite{wu2023topomlp}. Firstly, we pass the centerline coordinates to MLP and add them to the refined query features. Finally, we apply MLP to perform topology classification. 

    \subsection{Lane-Traffic topology}
    We use the coordinates of centerlines, and coordinates, categories from traffic element bboxes. We train topology model using ground truth data of lane segments and traffic elements, since the feature embedding are not used. By decoupling with the upstream detection model, the training and prediction process of topology becomes more convenient.
    Due to the complexity of intersections, we use self-attention to facilitate information exchange among elements and obtain relative relationships.

\section{Experiments}
\label{sec:experiments}

\subsection{Implementation details}
We build our system based on MapTRv2 codebase \cite{liao2023maptrv2}.
Training setup. We adopt two data augmentation methods, image data augmentation and BEV data augmentation, e.g. random rotating, scaling, cropping and flipping. For ablation study, we use the ResNet50 \cite{he2016deep} pretrained on ImageNet dataset. And we use ViT-L \cite{fang2023eva} as the scaling up image backbone. We pretrain the ViT model on nuScenes dataset with vectorized map construction task. For training large-scale models, we use a batch size of 16 on 16 A800 GPUs, AdamW \cite{loshchilov2017decoupled} optimizer with a learning rate of 6e-4. The layer-wise learning rate decay is 0.9. Partial freeze block number of ViT is 3. The resolution of input images are $1536\times1536$. And the image features from the backbone are downsampled with a stride of 16. The depth net predicts depth from 1m to 56m. The BEV feature-map resolution is $100\times200$.
We train the model by two stages. Single-frame mode for 48 epochs and streaming-stacking mode for 36 epochs. During the temporal fusion mode, we change the partial freeze block number of ViT to 12. And turn off both image and BEV data augmentations.

\subsection{Ablation Study}
    \subsubsection{SymmetricVT}
    We examine the efficacy of SVT component through ablation studies, utilizing the OpenlaneV2 dataset \cite{wang2024openlane}. Starting with BEVFormer \cite{li2022bevformer} and LSS \cite{philion2020lift} as baseline, the best score is 40.36\% on the validation set, as shown in Table \ref{tab:SVT}. Compared to the best baseline, the integration of BEVFormer and LSS increase 0.5\% mAP. After adding image data-augmentation and BEV data-augmentation, the model performance has improved to 43.75\%.

    \begin{table}
      \centering
      \begin{tabular}{@{}ccc|c@{}}
        \toprule
        BEVFormer & LSS & Data-aug & mAP\(\%\) \\
        \midrule
        \checkmark  &               &               & 40.36 \\
                    & \checkmark    &               & 32.57 \\
        \checkmark  & \checkmark    &               & 40.89 \\
        \checkmark  & \checkmark    & \checkmark    & 43.75 \\
        \bottomrule
      \end{tabular}
      \caption{Ablation study of SVT on the Openlanev2 val set.}
      \label{tab:SVT}
    \end{table}

    \subsubsection{Temporal fusion}
    We build a single-frame baseline model by training 72 epochs with ResNet50 checkpoints. And then all experiments finetune the baseline model by 12 epochs. We use single-frame mode to finetune the baseline model, model can reach a score of 52.93\% mAP, as shown in Table \ref{tab:HTF}. For the streaming strategy, we use one convGRU \cite{chung2014empirical} as dense fusion encoder. It has a performance improvement of 3.7\%. And for the streaming-streaming strategy, two layers of convGRU are used instead of one. Unfortunately, the performance increase only 0.56\% compared to single-frame. For the streaming-stacking mode, we select 4 frames out of latest 10 frames for the layer of stacking mode during training phase, and a certain distance strides of 5, 10, 15, 20 meters during testing phase. the performance reaches 57.13\% mAP.

    \begin{table}
      \centering
      \begin{tabular}{@{}l|c@{}}
        \toprule
        Temporal fusion strategy & mAP\(\%\) \\
        \midrule
        None & 52.93 \\
        Streaming & 56.61 \\
        Streaming-streaming & 53.49 \\
        Streaming-stacking & 57.13 \\

        \bottomrule
      \end{tabular}
      \caption{Ablation study of HTF on the Openlanev2 val set.}
      \label{tab:HTF}
    \end{table}

    \subsubsection{Ped crossing resampling}
    We use the hierarchical attention based decoder same as MapTR as baseline. The model performance reaches score of 33.6\% DET-a, as shown in Table \ref{tab:ped}. Then we change the decoder to instance attention. The model performance increase 0.45\%. Finally, we utilize ped crossing resampling to improve the performance to 35.42\%

    \begin{table}
      \centering
      \begin{tabular}{@{}l|c@{}}
        \toprule
        Method & DET-a\(\%\) \\
        \midrule
        ins-pt attention & 33.6 \\
        ins attention & 34.05 \\
        ins attention + ped crossing resampling & 35.42 \\

        \bottomrule
      \end{tabular}
      \caption{Ablation study of ped crossing resampling on the Openlanev2 val set. The ins-pt attention is short for hierarchical attention used in MapTR \cite{liao2022maptr} model.}
      \label{tab:ped}
    \end{table}

    \subsubsection{Traffic elements}
    We utilize COCO pretrained model and finetune for 40 epochs as our 2D detector baseline. The dataset is resampled by a ratio from 5 to 20 times. The entire model is optimized by AdamW with a learning rate of 0.04 and resolution of $1568\times2048$. And then we generate pseudo-labels by threshold of 0.3.
    YOLOv8-x with data augmentation can reach a score of 79.42\% on DET-l, as shown in Table \ref{tab:te}. The application of resampling has a performance improvement of 0.64\%. TTA further improves 1.0\% score. We utilize pseudo label to improve the performance to 81.81\%. Finally, the model ensemble of YOLOv8 and YOLOv9 \cite{wang2024yolov9} improve the performance to 82.4\%.

    \begin{table}
      \centering
      \begin{tabular}{@{}l|c@{}}
        \toprule
        Method & DET-t\(\%\) \\
        \midrule
        YOLOv8+data-aug & 79.42 \\
        +Resampling & 80.06 \\
        +TTA & 81.07 \\
        +Pseudo label learning & 81.81 \\
        +YOLOv9 ensemble & 82.40 \\
        \bottomrule
      \end{tabular}
      \caption{Ablation study of traffic elements on the Openlanev2 test set.}
      \label{tab:te}
    \end{table}

    \subsubsection{Lane segments}
    We train three versions of the models, using different backbones (ViT \cite{fang2023eva}, InternImage-XL \cite{wang2023internimage}) with different input image resolution scales (0.5, 0.75, 1). During the ensemble process, we utilize an ensemble strategy that incorporates predictions with low similarity. Initially, the models are sorted by their evaluation scores, the best model is the base model, and the other two models are subsequently integrated as proposal models.
    From the Table \ref{tab:ls}, it can be seen that the more models ensembled, the more remarkable performance improved.

    \begin{table}
      \centering
      \begin{tabular}{@{}ccc|ccc@{}}
        \toprule
        Model A & Model B & Model C & DET-l\(\%\) & TOP-ll\(\%\) & TOP-lt\(\%\) \\
        \midrule
        \checkmark  &               &               & 48.67 & 40.04 & 48.27 \\
                    & \checkmark    &               & 46.5  & 36.82 & 47.36 \\
                    &               & \checkmark    & 42.33 & 35.01 & 44.53 \\
        \checkmark  & \checkmark    &               & 49.8  & 43.0  & 50.97 \\
        \checkmark  & \checkmark    & \checkmark    & 50.74 & 46.32 & 53.59 \\
        \bottomrule
      \end{tabular}
      \caption{Ablation study of lane segments model ensemble on the Openlanev2 test set.}
      \label{tab:ls}
    \end{table}

\section{Conclusion}
\label{sec:conclusion}

In this work, we rethink the pipeline of 2D / 3D elements detection and topology reasoning of mapless driving. Firstly, we employ a symmetric view transformation(SVT) to combine forward projection and backward projection to form complementary advantages. Secondly, we introduce hierarchical temporal fusion(HTF) to integrate temporal features from local to global stably. Moreover, we improve ped crossing representation by a novel resampling method. Finally, LGmap is the first-place solution on the Mapless Driving track, which achieves 0.66 UniScore.

{\small
\bibliographystyle{unsrt}
\bibliography{11_references}
}

\ifarxiv \clearpage \appendix \section{Appendix Section}
\label{sec:appendix_section}
Supplementary material goes here.
 \fi

\end{document}


\title{\paperTitle}
\author{\authorBlock}
\maketitlesupplementary

\appendix
\section{Appendix Section}
\label{sec:appendix_section}
Supplementary material goes here.

{\small
\bibliographystyle{ieeenat_fullname}
\bibliography{11_references}
}